\documentclass[10pt,twocolumn,letterpaper]{article}

\usepackage{cvpr}
\usepackage{times}
\usepackage{epsfig}
\usepackage{graphicx,subfigure}
\usepackage{amsmath}
\usepackage{amssymb}
\usepackage{bm}
\usepackage{multirow}

\usepackage{enumitem}

\usepackage{array}

\newcommand{\rulesep}{\unskip\ \vrule\ }

\usepackage[font=small,labelsep=period]{caption}
\newcommand{\myparagraph}[1]{\smallskip\noindent\textbf{#1.}}
\def\F{\mathcal{F}}

\def\M{\mathcal{M}}

\def\X{\mathcal{X}}
\def\x{\textbf{x}}
\def\z{\textbf{z}}

\newcommand{\mcb}{\color{black}}

\newcolumntype{L}[1]{>{\raggedright\let\newline\\\arraybackslash\hspace{0pt}}m{#1}}
\newcolumntype{C}[1]{>{\centering\let\newline\\\arraybackslash\hspace{0pt}}m{#1}}
\newcolumntype{R}[1]{>{\raggedleft\let\newline\\\arraybackslash\hspace{0pt}}m{#1}}

\usepackage[pagebackref=true,breaklinks=true,letterpaper=true,colorlinks,bookmarks=false]{hyperref}

\cvprfinalcopy 

\def\cvprPaperID{460} 

\ifcvprfinal\fi
\begin{document}

\title{Dual Attention Matching Network for Context-Aware Feature Sequence based Person Re-Identification}

\author{
Jianlou Si\textsuperscript{1}, Honggang Zhang\textsuperscript{1}, Chun-Guang Li\textsuperscript{1}, Jason Kuen\textsuperscript{2}, \\Xiangfei Kong\textsuperscript{2}, Alex C. Kot\textsuperscript{2}, Gang Wang\textsuperscript{3}\\
\textsuperscript{1} Beijing University of Posts and Telecommunications, Beijing, China\\
\textsuperscript{2} Nanyang Technological University, Singapore\\
\textsuperscript{3} Alibaba AI Labs, Hangzhou, China\\
}

\maketitle

\begin{abstract}
{\mcb
}


Typical person re-identification (ReID) methods usually describe each pedestrian with a single feature vector and match them in a task-specific metric space. However, the methods based on a single feature vector are not sufficient enough to overcome visual ambiguity, which frequently occurs in real scenario. In this paper, we propose a novel end-to-end trainable framework, called Dual ATtention Matching network (DuATM), to learn context-aware feature sequences and perform attentive sequence comparison simultaneously. The core component of our DuATM framework is a dual attention mechanism, in which both intra-sequence and inter-sequence attention strategies are used for feature refinement and feature-pair alignment, respectively. Thus, detailed visual cues contained in the intermediate feature sequences can be automatically exploited and properly compared. We train the proposed DuATM network as a siamese network via a triplet loss assisted with a de-correlation loss and a cross-entropy loss. We conduct extensive experiments on both image and video based ReID benchmark datasets. Experimental results demonstrate the significant advantages of our approach compared to the state-of-the-art methods.


\end{abstract}

\section{Introduction}

\begin{figure}[t]
\begin{center}
   \includegraphics[width=0.9\linewidth]{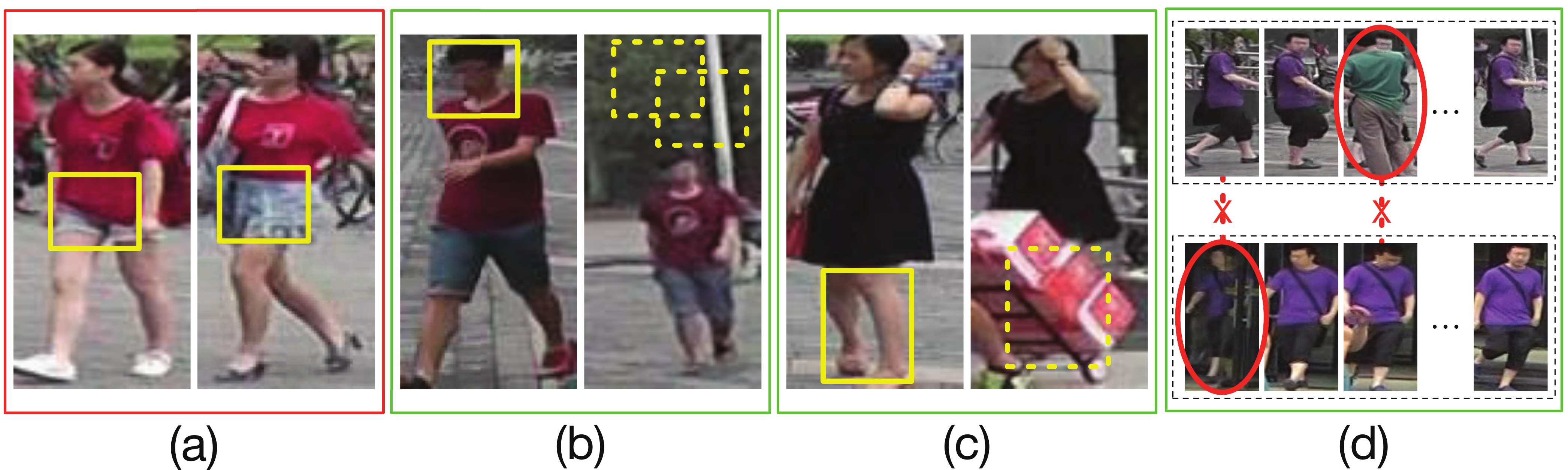}
\end{center}
   \caption{{\mcb Hard examples in person ReID. (a): negative pair with similar appearance. (b): positive pair with large spatial displacement. (c):  positive pair with body-part missing due to occlusion. (d): positive video pair with interference frames (marked by ellipse) and temporal misalignments (indicated by red ``$\times$'' mark).}}
\label{fig:example}
\vspace{-3mm}
\end{figure}

Person Re-Identification (ReID) aims at associating the same pedestrian across multiple cameras \cite{Gong2011Person,Zheng2016Person}, which has attracted rapidly increased attention in the computer vision community due to its importance for many potential applications, such as video surveillance analysis and content-based image/video retrieval.
%
%
A typical {person ReID pipeline} usually describes each pedestrian image or video footage with \textit{a single feature vector} firstly and then match them in a task-specific metric space, where the feature vectors of same pedestrian are expected to have smaller distances than that of different pedestrians, \eg, \cite{Li2013Learning, Si:ICIP15, Liao2015Person,Zheng2013Reidentification,You2016Top}. Recently, benefited from the success of deep learning, 
feature vector based methods have obtained significant performance improvements~\cite{Hermans2017Defense, Sun2017Svdnet, Mclaughlin2016Recurrent, Liu2017Quality}. However, when the individuals undergo drastic appearance changes or when they are dressed in similar clothes, {it becomes quite difficult to use single feature vector based representation for reliable person ReID}. As shown in Fig.~\ref{fig:example}~(a), different individuals are very similar to each other in appearance, except for some local patterns on skirts. Unfortunately, the single feature vector based methods usually pay more attention on the overall appearance rather than the local discriminative parts and thus fail to yield accurate matching results. Moreover, as shown in Fig.~\ref{fig:example}~(d), there are also some interference frames in each video sequence, which will heavily contaminate the whole feature vector 
and thus lead to mismatching.

An alternative way to address these problems is to describe each person with \textit{a set of feature vectors} and match them based on feature set or feature sequence.\footnote{In this paper, we refer to a group of feature vectors as feature sequence if they have spatial/temporal adjacent relations; otherwise as feature set.} For example, in \cite{Li2014Deepreid, Ahmed2015Improved, Shen2015Person, Zhao2017Spindle, Zhao2017Deeply}, the spatial-patch based local feature sequences or body-part based semantic feature sets are extracted from pedestrian images and matched according to some heuristic correspondence structures; in \cite{Wang2016Person, Wang2014Person, Zhou2017See}, multiple sub-segment or frame level comparisons are computed and aggregated for matching pedestrian based on video.
Among these methods, matching the local feature sequences based on a universal correspondence structure might easily fail when encountering heavily sequence misalignments, \eg, caused by the spatial displacements as shown in Fig.~\ref{fig:example}~(b) or local interferences as shown in Fig.~\ref{fig:example}~(d). Besides, matching the semantic feature sets based on human body structure might also fail when encountering body-part occlusions 
as shown in Fig.~\ref{fig:example}~(c). 

\begin{figure}[t]
\begin{center}
   \includegraphics[width=0.9\linewidth]{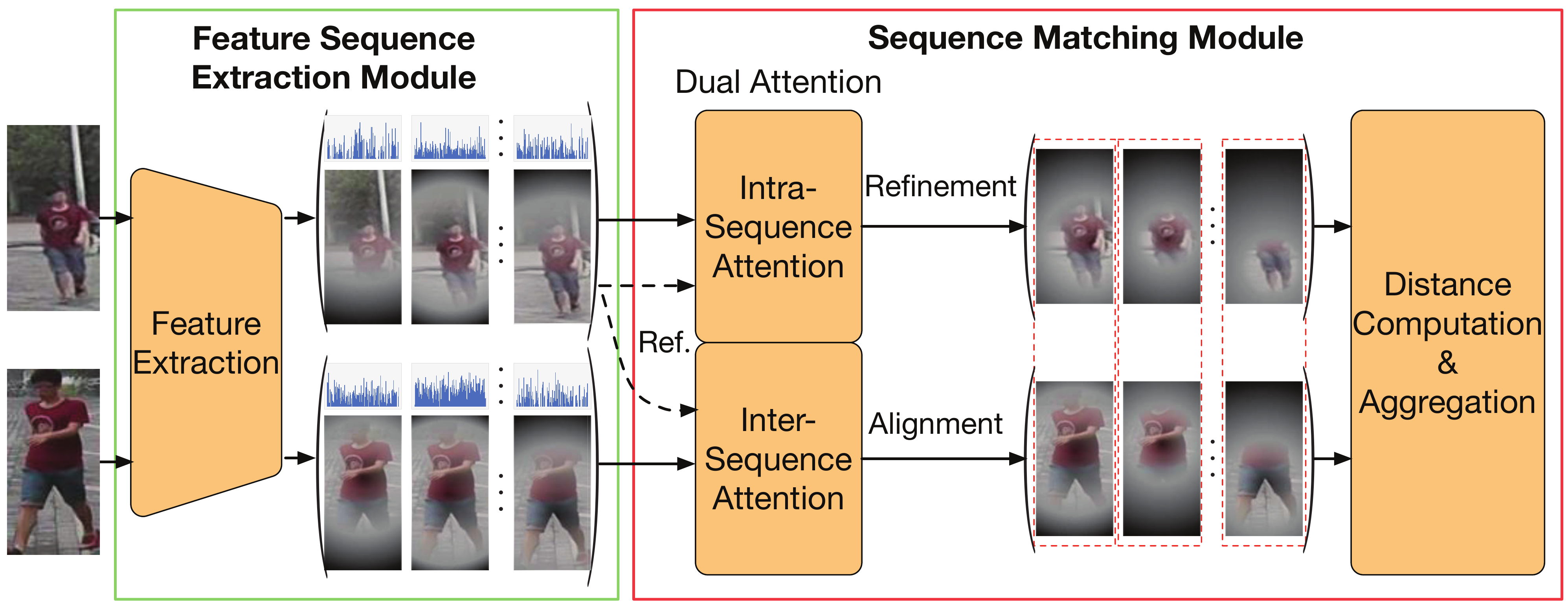}
\end{center}
   \caption{Schematic illustration of DuATM.} 
\label{fig:framenew}
\vspace{-3mm}
\end{figure}

To tackle the challenges mentioned above, we propose a novel end-to-end trainable framework, named Dual ATtention Matching network (DuATM), to jointly learn context-aware feature sequences and perform attentive sequences comparison. Our framework consists of two cascaded modules, one for feature sequence extraction and one for feature sequence matching, as illustrated in Fig.~\ref{fig:framenew}.
The feature sequence extraction module is built on a spatial/bi-recurrent convolutional neural network (CNN) for image/video inputs to extract spatial/temporal-spatial context-aware feature sequences. 
The sequence matching module is based on a dual attention mechanism---which contains one attention strategy for intra-sequence refinement and one attention strategy for inter-sequence alignment---the former refines each corrupted feature vectors by exploiting the contextual information within sequence and the later aligns feature-pair by selecting semantically consistent counterparts cross paired sequences.
After feature sequences refinement and alignment, the holistic sequence distance score is computed by aggregating multiple local distances between the refined and aligned pairwise feature vectors of each paired sequences.
We train the proposed DuATM as a siamese network with a triplet loss 
plus a de-correlation loss and a cross-entropy loss, to reduce the feature sequence redundancy and enhance the feature sequence discrimination. 

The main contributions of the paper are as follows.
\begin{itemize}[leftmargin=*]
\item We propose a novel end-to-end trainable framework for person ReID, which can jointly learn context-aware feature sequences and perform sequences comparison with dual attention mechanism.

\item We use a dual attention mechanism to perform intra-sequence feature refinement and inter-sequence feature-pair alignment simultaneously.

\item We train DuATM as a siamese network with a triplet loss, plus a de-correlation loss and a cross-entropy loss, and evaluate the effectiveness of each part.

\item We conduct extensive experiments on both image and video based benchmark datasets and demonstrate the effectiveness of our proposal.
\end{itemize}

{\mcb

\section{Related Works}

Person ReID systems usually consist of two major components: a) feature extraction and b) metric learning. Previous works on person ReID focus on either constructing informative features, or finding a discriminative distance metric. 
%
According to the used representation forms in matching stage, we roughly divide the existing methods into two groups:
\textit{feature vector based methods}, \eg, \cite{Cheng2017Discriminative,Chen2017Fast,Li2017Person,Sun2017Svdnet,Wang2016Joint,Li2017Learning,Geng2016Deep,Xiao2016Learning,Si:TSMC17}; and \textit{feature set or feature sequence based methods}, \eg, \cite{Zhou2017See,Zhao2017Deeply,Zhao2017Spindle,Zhang2016Semantics,Subramaniam2016Deep,Li2014Deepreid,Ahmed2015Improved,Varior2016Gated}.

In feature vector based methods, an image or video is represented by a feature vector and the metric learning is performed based on feature vectors. For example, in~\cite{Bai2017Scalable, Li2013Learning, Mignon2012Pcca, Xiong2014Person, Zhang2016Sample, Yang2016Large, You2016Top, Zheng2013Reidentification, Zhu2016Video, Liao2015Person}, hand-crafted local features are integrated into a feature vector, and distance metric is learned by simultaneously maximizing inter-class margins and minimizing intra-class variations. Meanwhile, many recent works directly learn comparable feature embedding from the raw input data via a neural network. For example, in \cite{Xu2017Jointly, Liu2017Quality}, high-quality local patterns are explored from images or videos firstly and then aggregated into informative feature vectors; in \cite{Mclaughlin2016Recurrent, Varior2016Siamese, Yan2016Person}, local features of recurrent appearance data are extracted and integrated using temporal-pooling strategy; in \cite{Hermans2017Defense, Chen2017Beyond}, to enhance the generalization capability of the learned embeddings, the pairwise similarity criterion is extended to triplet or quadruplet. Although these methods mentioned above are able to learn task-specific compact embeddings, these methods still suffer from the mismatching problem, especially when some vital visual details fail to be captured. 

Different from the feature vector based methods, feature set or feature sequence based methods are capable of preserving more detailed visual cues by leveraging complementary feature vectors or spatial information. 
For example, in \cite{Shen2015Person, Chen2016Similarity}, local spatial constraints are adopted when computing spatial-patch based feature sets similarity;
in \cite{Wang2014Person}, dense element-wise correspondences are employed when computing the distance of temporal feature sequences;
in \cite{Subramaniam2016Deep, Li2014Deepreid, Ahmed2015Improved, Varior2016Gated}, spatial correspondence structures are explored via the patch comparison layer inserted in a deep network;
in \cite{Zhao2017Spindle, Zhao2017Deeply}, the body structure information is utilized to facilitate the semantic alignment of feature sequences.
While these methods mentioned above exploit heuristic correspondence structures to compare feature sequences, they might easily fail when heavy misalignments or interferences occur in feature sequences. 

Recently, attention mechanism has been proposed 
in many tasks of matching sequences or learning representations, \eg,~\cite{Vinyals2015Order, Xu2015Show, Yang2017Neural,
Wang2017Compare, Liu2017End, Zhou2017See}. In \cite{Vinyals2015Order,Wang2017Compare}, attention mechanism is used to softly align word embeddings between each text-sequence pair in the task of natural language processing. In \cite{Xu2015Show,Liu2017End}, glimpse representation is produced for each image via neural attention, so that each input pair can be compared progressively. In \cite{Yang2017Neural,Zhou2017See}, fixed-dimension feature vectors are learned from variable length videos for face recognition and person ReID by attentive aggregation, respectively.
However, these methods either consider a single intra-sequence attention for feature selection from feature set/sequence,
or consider a single inter-sequence attention for feature sets/sequences matching.
While inter-sequence attention is able to tackle the sequence misalignment problem, it might fail when interferences or corruptions occur. On the other hand, intra-sequence attention is able to tackle corruptions but it is not able to align sequences.

In this paper, we exploit the attention mechanism into feature sequence based person ReID.  
Unlike the existing methods, that compare sequences via heuristic correspondence structures, we attempt to compare two sequences via \textit{dual attention processes}, in which an inter-sequence attention process is used to perform sequence alignment
and an intra-sequence attention process is used simultaneously to perform sequence refinement.

{\mcb
\section{Our Proposal: Dual Attention Matching Network (DuATM)}

This section will present an end-to-end trainable framework---DuATM, which consists of two modules: one for extracting feature sequences and one for matching feature sequences, as illustrated in Fig.~\ref{fig:framenew}. 


\begin{figure}[t]
\vspace{-0mm}
\centering
\subfigure[Image feature extraction]{\includegraphics[clip=true,trim=0 0 0 0,width=0.465\columnwidth]{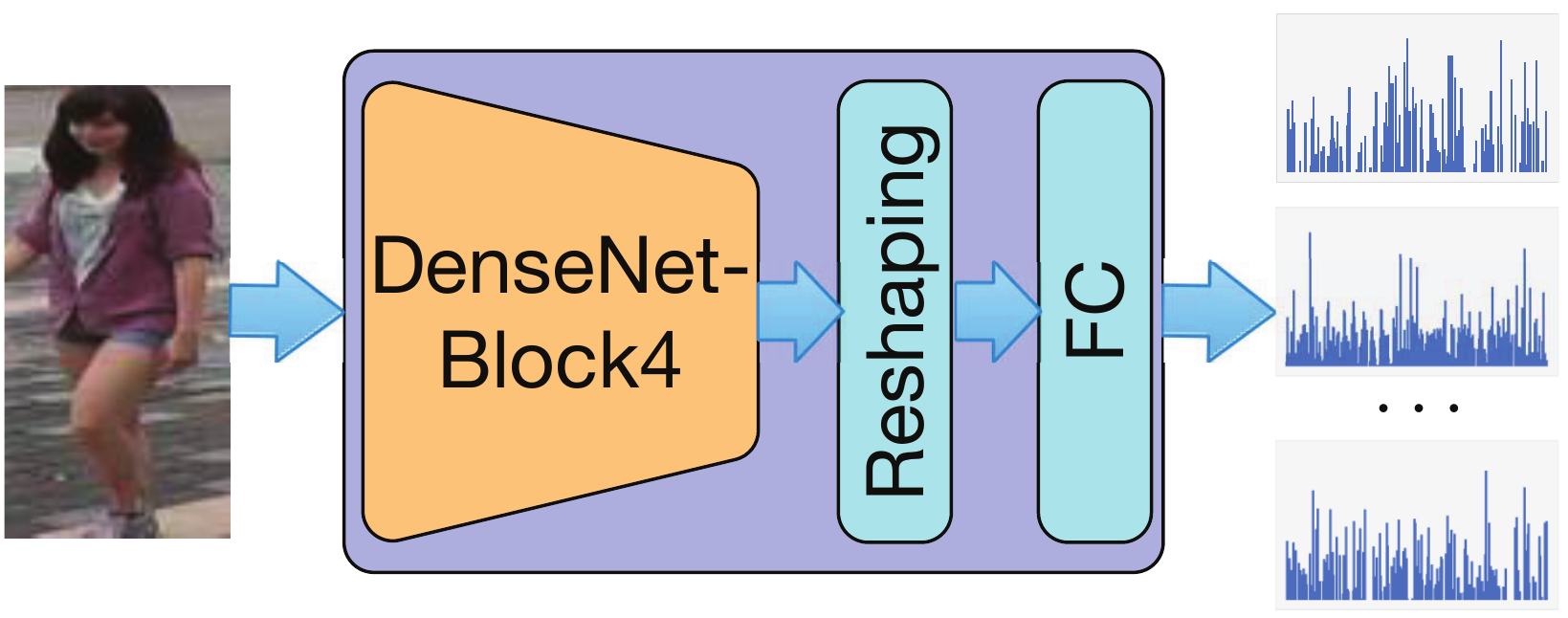}}
\rulesep
\subfigure[Video feature extraction]{\includegraphics[clip=true,trim=0 0 0 0,width=0.465\columnwidth]{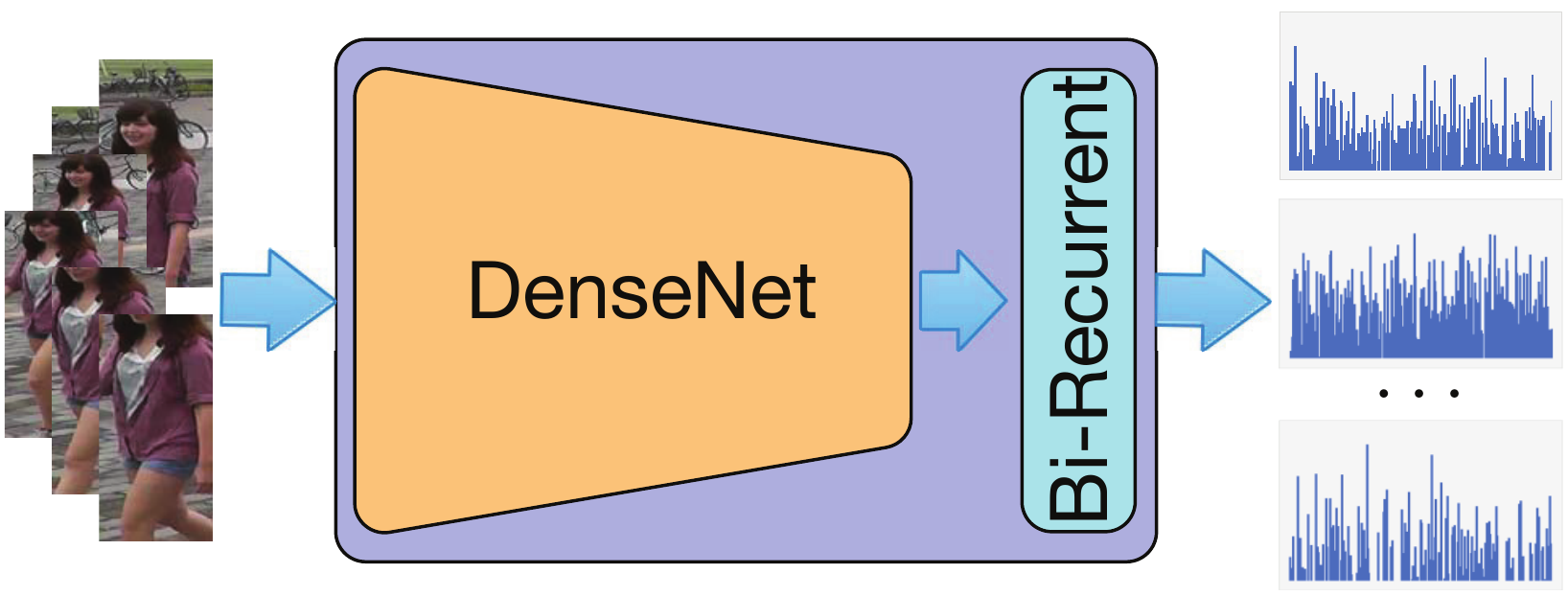}}
\caption{Feature sequence extraction module.}
\label{fig:feat-module}
\vspace{-3mm}
\end{figure}

\subsection{Feature Sequence Extraction Module}


In DuATM, we adopt DenseNet-121 \cite{Huang2017Densely} as the backbone of the feature sequence extraction module. Owning to the direct connections between each layer to all the subsequent layers in DenseNet, local details are better propagated to the outputs to enrich the final feature sequences. Specifically, the network architectures for image and video inputs are slightly different. 

\begin{itemize}[leftmargin=*]
\item Given an image $\mathcal{X} \in \mathbb{R}^{H \times W\times 3}$, as illustrated in Fig.~\ref{fig:feat-module} (a), the convolutional feature maps are obtained from the outputs of DenseNet-Block4. 
    Each feature vector at a specific position across all channels contains both the spatial details and semantic contexts due to its large receptive field size. Then, these feature vectors are rearranged by locations to form a feature sequence and each feature vector is further transformed into a compact embedding space via a Fully Connected (FC) layer.

\item Given a video footage $\mathcal{X} \in  \mathbb{R}^{H \times W \times 3 \times T}$, of length $T$, as illustrated in  Fig.~\ref{fig:feat-module} (b), each frame in video at a time-step is passed to a DenseNet to produce the frame-level feature vector. Then, a bidirectional recurrent layer is introduced to encode both the temporal-spatial appearance details and the complementary motion cues, by allowing information to be passed between time-steps. 
    Finally, all hidden states from different time-steps compose the final feature sequence for the video.
\end{itemize}


For convenience, we denote the feature sequence extraction as $\mathbf{X} = \F(\X; \Theta_{\F})$, where $\mathbf{X}$ is the extracted feature vectors sequence which encodes spatial or temporal information and $\mathcal{F}(\cdot; \Theta_{\F})$ represents the feature extraction module parameterized with $\Theta_{\F}$. 
More specifically, we denote $\mathbf{X}=\{\mathbf{x}^{i} \in \mathbb{R}^D\}_{i=1}^S$ as a feature sequence of length $S$. 
Each feature vector $\mathbf{x}^{i}$ is normalized to unit $\ell_2$ norm before passing it to the next module.

\subsection{Sequence Matching Module}

\begin{figure*}[t]
\vspace{-0mm}
\centering
\subfigure[Sequence matching module]{\includegraphics[clip=true,trim=0 0 5 0,width=1.00\columnwidth]{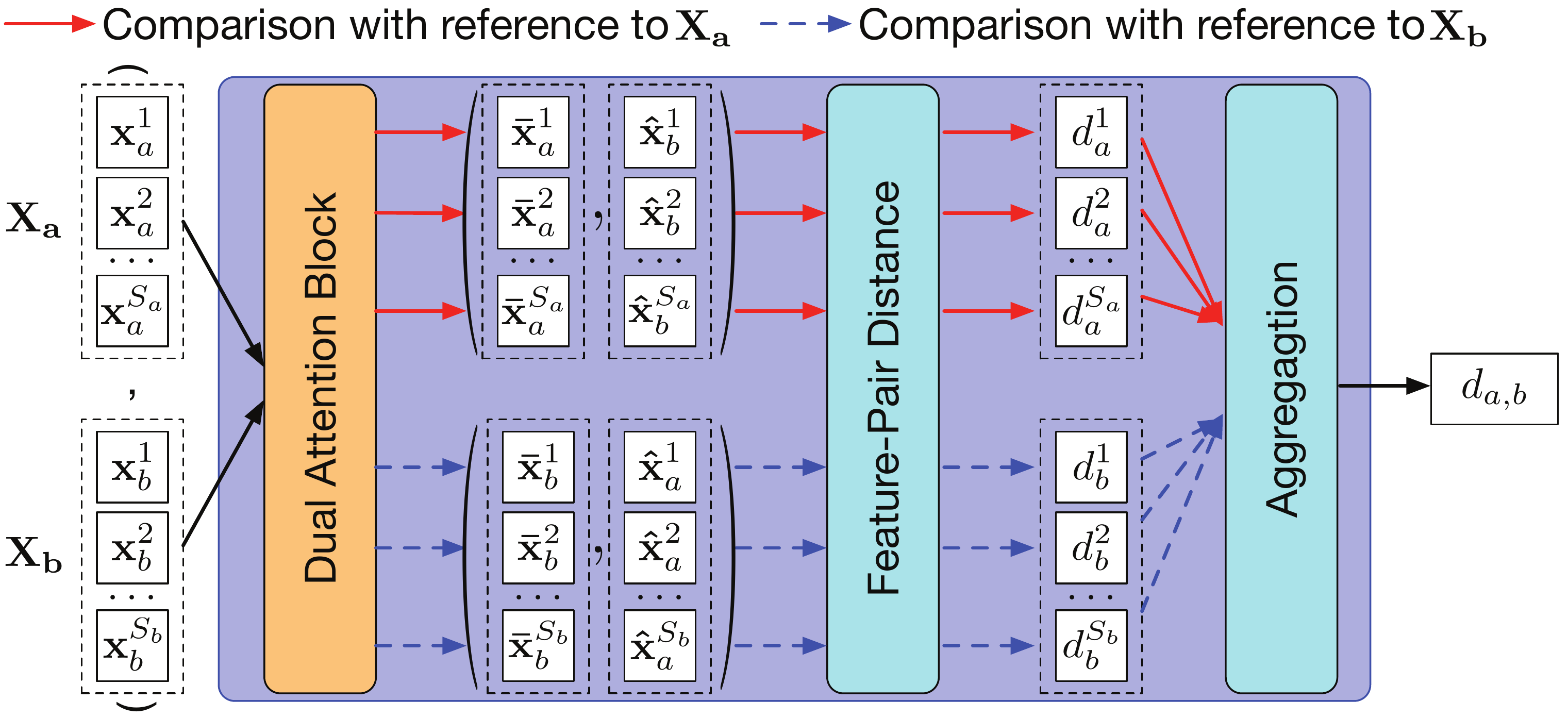}}
\quad
\quad
\subfigure[Dual attention block]{\includegraphics[clip=true,trim=0 0 0 0,width=0.87\columnwidth]{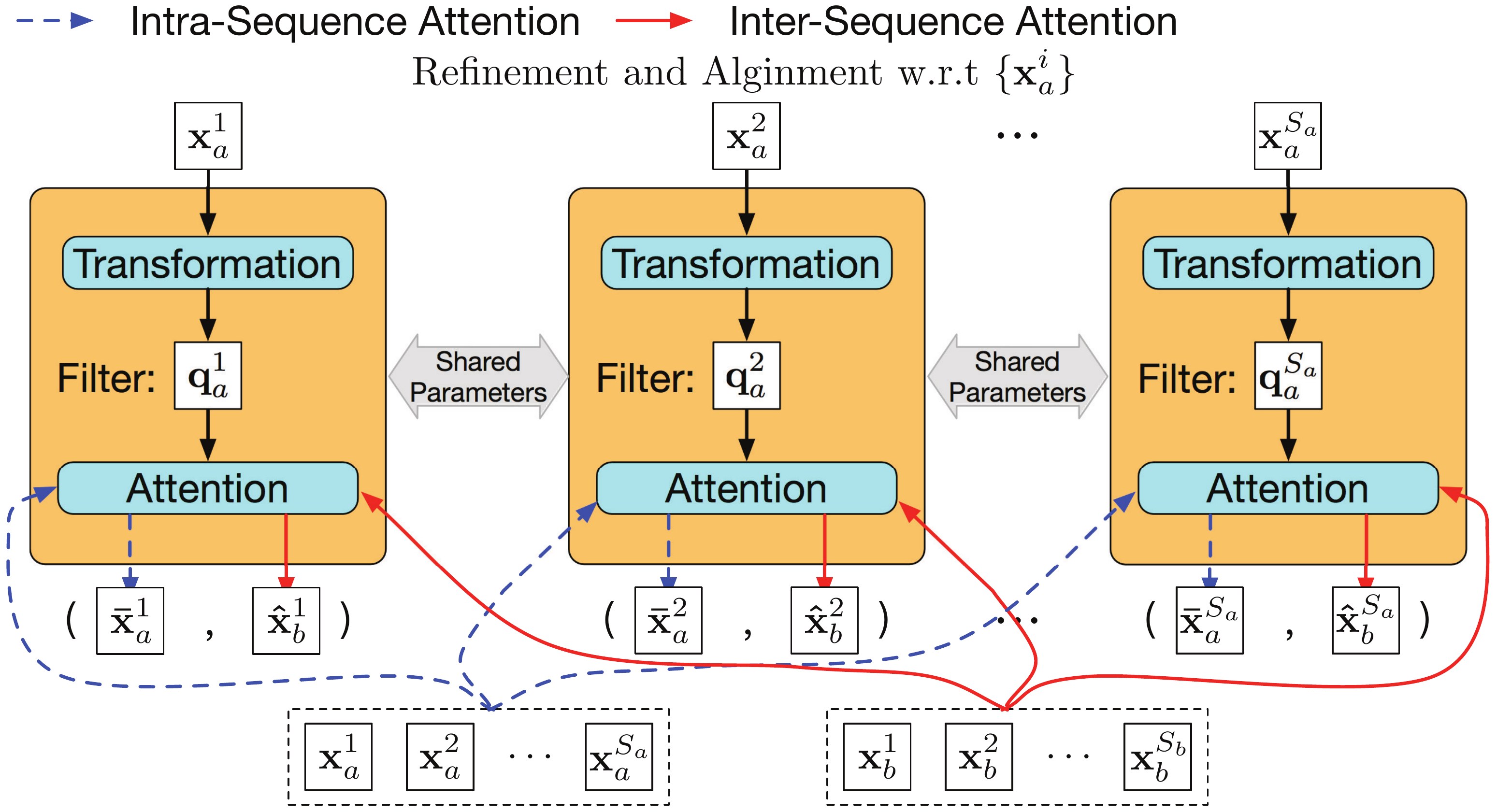}}
\caption{Illustration for sequence matching module and dual attention block.}
\label{fig:match-module}
\vspace{-3mm}
\end{figure*}


Sequence matching module is the most important component of DuATM. Note that there is no supervision information available to force the feature extraction module to learn semantically aligned feature sequences, thus one of the goals of this module is to compare each pair of possibly unaligned feature sequences $(\mathbf{X}_a, \mathbf{X}_b)$, where $\mathbf{X}_a = \{\mathbf{x}_a^i\}_{i=1}^{S_a}$ and $\mathbf{X}_b = \{\mathbf{x}_b^j\}_{j=1}^{S_b}$. However, each sequence may also contain a certain amount of corrupted feature vectors (\eg, caused by the noisy inputs). A naive 
method is to transform feature sequences into comparable 
feature vectors via average pooling, in which the misalignment or corruptions are ignored. 
Instead, we propose to refine and align each feature sequence pair at first, then compute and aggregate the distances of multiple feature pairs.

Since that the intermediate feature sequences obtained from our feature extraction module contain abundant contextual information, we 
use these contexts 
to remove the feature corruptions and 
compare feature sequences. 
Specifically, we attempt to exploit the contextual information to help feature sequence refinement and feature sequence pair alignment via the attention mechanism. 
To be more specific, if one of a feature sequence pair is treated as the memory, the refinement of each feature vector within this sequence can be achieved by an intra-sequence attention process; oppositely, if the other sequence is treated as the memory, the alignment of each feature vector can be achieved by an inter-sequence attention process; and vice versa. 

For clarity, we illustrate the sequence matching module of DuATM in Fig.~\ref{fig:match-module}~(a), in which two types of attention procedures are integrated into a dual attention block for feature sequence refinement and alignment, as illustrated in Fig.~\ref{fig:match-module} (b). After refinement and alignment, the holistic sequence distance score is computed by aggregating multiple local distances between refined and aligned pairwise feature vectors of each sequence pair.

\subsubsection{Dual Attention Block}

The dual attention block is composed of one transform layer and one attention layer, in which the transform layer is used to produce the feature-aware filter and the attention layer is used to generate the corresponding attention weights.
Without loss of generality, as an example to present the dual attention block in detail, we describe the generation process of $(\mathbf{\bar{x}}_a^i,\mathbf{\hat{x}}_b^i)$, as illustrated in Fig. \ref{fig:match-module} (b). Specifically, let $\mathbf{x}_a^i$ be the reference feature to be refined and aligned.
\begin{itemize}[leftmargin=*]
\item At first, the filter is computed through the transform layer as follows:
\begin{equation}
\mathbf{q}_a^i  =  ReLU(BN(\mathbf{W} \mathbf{x}_a^i + \mathbf{b})),
\label{eq:eq1}
\end{equation}
where $\mathbf{W}$ and $\mathbf{b}$ are the weight matrix and bias vector of a linear layer, $BN$ and $ReLU$ represent Batch Normalization \cite{Ioffe2015Batch} and rectified linear unit (ReLU) function, respectively.
\item Then, the attention significance for intra-sequence refinement and inter-sequence alignment can be computed separately through the attention layer as follows:
\begin{equation}
\bar{e}_a^{i,m} = \langle{\mathbf{q}_a^i}, \mathbf{x}_a^m\rangle, ~~~~ \hat{e}_b^{i,n} = \langle{\mathbf{q}_a^i} ,  \mathbf{x}_b^n\rangle,
\label{eq:eq2}
\end{equation}
where $\langle\cdot,\cdot\rangle$ denotes the inner product.
\item Finally, the semantically refined and aligned feature vector pair $(\mathbf{\bar{x}}_a^i,\mathbf{\hat{x}}_b^i)$ is obtained by linearly combining elements within the corresponding sequences, respectively, via normalized attention weights as
%
\begin{equation}
\mathbf{\bar{x}}_a^i = \sum_{m=1}^{S_a}{\sigma(\bar{e}_a^{i,m}){\mathbf{x}_a^m}}, ~~~ \mathbf{\hat{x}}_b^i = \sum_{n=1}^{S_b}{\sigma(\hat{e}_b^{i,n}){\mathbf{x}_b^n}},
\label{eq:eq3}
\end{equation}
where $\sigma(\cdot)$ is a soft-max function for normalization in which $\sigma(t_j)=\frac{\exp(t_j)}{\sum_{j=1}^S \exp(t_j)}$ for $\mathbf{t} \in \mathbb{R}^S$.

\end{itemize}

Following 
Eq.~\eqref{eq:eq1} to \eqref{eq:eq3}, the 
comparable feature pairs with respect to each feature vector 
can be obtained.

\subsubsection{Distance Computation and Aggregation}


Owing to the feature sequence refinement and alignment performed in the dual attention block, it 
is reasonable to directly compute the distance between two refined and simultaneously aligned features, and aggregate the computed distances of feature-pairs into a holistic sequence distance.

In DuATM, the dual attention is bidirectional, \ie, the dual attention process is carried out twice with respect to $\{\mathbf{x}_a^i\}$ and $\{\mathbf{x}_b^j\}$, respectively. Thus, we use the the distances of sequence-pair in both two comparison directions 
to define the distance of the holistic sequences.
%
Specifically, we use the Euclidean distance to compute the distance between feature pair, \ie,
\begin{equation}
\begin{split}
d_a^i = \| \bar \x_a^i - \hat \x_b^i \|_2,~~~i=1,\cdots,S_a,\\
d_b^j = \| \bar \x_b^j - \hat \x_a^j \|_2,~~~j=1,\cdots,S_b.
\end{split}
\label{eq:seq-pair-distance}
\end{equation}
And then, we aggregate these distances via the average-pooling to define the distance of feature sequences $\mathbf{X}_a$ and $\mathbf{X}_b$ as follows:
\begin{equation}
\|\mathbf{X}_a - \mathbf{X}_b \|_{_\M} = \frac{1}{2S_a}\sum_{i=1}^{S_a} d_a^i + \frac{1}{2S_b}\sum_{j=1}^{S_b} d_b^j,
\label{eq:seq-pair-distances-aggregation}
\end{equation}
where $\|\mathbf{X}_a - \mathbf{X}_b \|_{_\M}$ is the distance defined by the sequence matching module.
%
For convenience, we denote all parameters (\ie, $\mathbf{W}$, $\mathbf{b}$, and the parameters in the BN layer) in the sequence matching module as $\Theta_{_\M}$.



\subsection{Loss Functions for Training DuATM}

 \begin{figure}[t]
\begin{center}
   \includegraphics[width=0.9\linewidth]{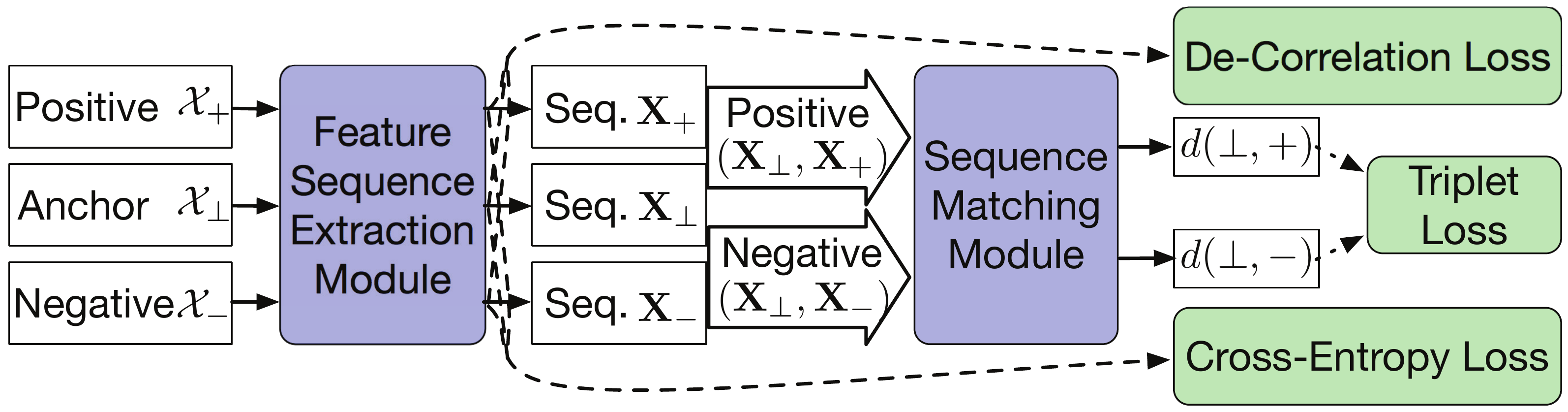}
\end{center}
\caption{Overall flowchart for training.}
\label{fig:train-module}
\vspace{-3mm}
\end{figure}

To train the whole network to perform person ReID and generalize well on unseen data, we use the siamese architecture with \textit{triplet loss} during the training period as shown in Fig. \ref{fig:train-module}. Moreover, 
to make the learned intermediate feature sequences 
compact, robust, and more discriminative, we also combined with two auxiliary losses, \ie \textit{de-correlation loss} and \textit{cross-entropy loss}. Thus, the overall loss function is defined as:
\begin{equation}
\ell = \ell^{(0)}(\X, \Theta_\F,\Theta_\M) + \lambda_1 \ell^{(1)}(\X, \Theta_\F) + \lambda_2 \ell^{(2)}(\X, \Theta_\F, \bm{\theta}),\nonumber
\end{equation}
where $\lambda_1>0$ and $\lambda_2>0$ are two tradeoff parameters. 


\myparagraph{Triplet Loss} The 
objective of using triplet loss is to force 
the network to make the distance between the positive pairs smaller than the negative ones.
Given a triplet of person images/videos, the extraction module extracts spatial/temporal-spatial context-aware feature sequences via a three-branch siamese subnet, and the matching module attentively computes the distances between the positive and negative pair via a two-branch siamese subnet.

Let $\X =(\mathcal{X}_{\bot}, \mathcal{X}_{+}, \mathcal{X}_{-})$ be a triplet input. To force the network to predict the distance of  positive pair smaller than the negative pair with a 
margin $\gamma$, we define the triplet loss $\ell^{(0)}(\X, \Theta_\F, \Theta_\M)$ as: 
\begin{equation}
\max\{0, \gamma + \|\F(\mathcal{X}_{\bot})-\F(\mathcal{X}_{+})\|_{_\M} -\|\F(\mathcal{X}_{\bot}),\F(\mathcal{X}_{-})\|_{_\M}\},
\end{equation}
where $\gamma > 0$ (\eg, $\gamma=0.2$ in our experiments).

\myparagraph{De-Correlation Loss}
In \cite{Cogswell2016Reducing}, de-correlating representations has been studied as a regularizer to reduce over-fitting in deep networks.
In this paper, we formulate a similar but different de-correlation loss to make our feature sequence more compact. 
Specifically, we 
impose a constraint on the intra-sequence correlation matrix as follows:
\begin{equation}
\ell^{(1)}(\X, \Theta_\F) = \frac{1}{N^2}||\mathbf{I} - \F(\mathcal{X})^T \F(\mathcal{X})||_F^2,
\end{equation}
where $\mathbf{I}$ is an identity matrix and $N$ is the total number of feature vectors in the sequence.

\myparagraph{Cross-Entropy Loss with Data Augmentation} To learn more informative and robust feature sequences, we also use a cross-entropy loss with data augmentation approach. Specifically, we use the data with the same labels to generate more data for training.

Suppose $\mathbf{X}=\F(\mathcal{X})$ is an intermediate feature sequence, we achieve this goal by first pooling the sequence as $\z=\sum_{i=1}^{S}{\omega_i \mathbf{x}_i}$, where $\sum_i \omega_i=1$ and $\omega_i \ge 0$, and then passing the aggregated vector to a FC layer followed by a cross-entropy loss:
\begin{equation}
\ell^{(2)}(\X, \Theta_\F, \bm{\theta})=-\ln{\sigma{{(\mathbf{w}_c} \z + b_c)}},
\end{equation}
where $c$ is the same label as the input $\mathcal{X}$, $\{\mathbf{w}_c, b_c\}$ refer to the $c_{\text{th}}$ row of the FC layer's weight matrix and bias vector, respectively, and $\bm{\theta}$ contains the parameters in the new FC layer. Note that, instead of generating $\z$ by simply average pooling, we propose to introduce a random \textit{convex combination} strategy into the pooling stage by randomly generating $\omega_i\in [0,1]$ and even reset it to 0 with the probability $p>0$, but keeping $\sum_i \omega_i = 1$. This can be regarded as a simplified version of the interpolation method \cite{DeVries2017Dataset} to augment training dataset.

\section{Experiments}

\begin{table*}
\small
\begin{center}
\begin{tabular}{ l | c c c c | c c c c | c c c c}
\hline
\multirow{2}{*}{Method \& Loss}
& \multicolumn{4}{c}{Market-1501}  & \multicolumn{4}{|c}{DukeMTMC-reID} & \multicolumn{4}{|c}{MARS}\\
\cline{2-13}
&  R1 & R5 & R20 & mAP & R1 & R5 & R20 & mAP & R1 & R5  & R20 & mAP \\
\hline
\hline
AvePool+$\ell^{(0)}$ 	& 74.20 & 89.67 & 95.58 & 56.88 & 64.05 & 79.44 & 87.52 & 43.79 & 65.45 & 81.92 & 90.10 & 47.26 \\
\hline 
DuATM+$\ell^{(0)}$ 	             	& 79.66 & 91.15 & 96.73 & 63.46 & 68.40 & 81.73 & 89.77 & 48.65 & 66.36 & 83.13 & 90.40 & 48.44\\
DuATM+$\ell^{(0)}$+$\ell^{(1)}$ 		& 81.83 & 92.46 & 97.33 & 65.21  & 69.17 & 82.23 & 89.36 & 49.48 & 66.52 & 83.78 & 91.21 & 49.07\\
DuATM+$\ell^{(0)}$+$\ell^{(2)}$   	& 87.50 & 95.37 & 98.01 & 70.02 & 79.40 & 90.04 & 94.25 & 61.55 & 73.74 & 87.73 & 93.84 & 56.36\\
DuATM+$\ell^{(0)}$+$\ell^{(1)}$+$\ell^{(2)}$ & 88.75 & 95.78 & 98.46 & 70.46 & 81.06 & \textbf{91.11} & 95.02 & 62.27 & 74.43 & 89.08 & 94.13 & 58.19\\
\hline 
DuATM$^*$+$\ell^{(0)}$+$\ell^{(1)}$+$\ell^{(2)}$        & 89.96 & 96.53 & 98.72 & 75.22 & 81.46 & 90.75 & 95.11 & 63.14 & 76.36 & 90.10 & 95.30 & 58.96\\
DuATM$^{**}$+$\ell^{(0)}$+$\ell^{(1)}$+$\ell^{(2)}$	& \textbf{91.42} & \textbf{97.09} & \textbf{98.96} & \textbf{76.62} & \textbf{81.82} & 90.17 & \textbf{95.38} & \textbf{64.58}  & \textbf{78.74} 	&\textbf{ 90.86}	& \textbf{95.76}	& \textbf{62.26}\\
\hline
\end{tabular}
\end{center}
\caption{Comparison to the baseline model. $^*$ We adjust the parameters of loss functions to more appropriate values as obtained in the parameter analysis experiments. $^{**}$ The data augmentation is also adopted during the evaluation stage.}
\label{tab:baseline}
\vspace{-3mm}
\end{table*}

To evaluate our proposal, we conduct extensive experiments on three large-scale data sets, including {Market-1501} \cite{Zheng2015Scalable}, {DukeMTMC-reID} \cite{Zheng2017Unlabeled}, and {MARS} \cite{Zheng2016Mars}.

\subsection{Datasets, Evaluation, and Implementations}

\myparagraph{Datasets Description} Market-1501 is collected from 6 cameras, which contains totally 1,501 identities and 32,668 bounding boxes generated by a DPM-detector. It is split into non-overlapping  train/test sets of 12,936/19,732 images as defined in \cite{Zheng2015Scalable}, and {single-query evaluation mode is adopted in our experiments}. DukeMTMC-reID is a subset of DukeMTMC \cite{Ristani2016MTMC} captured with 8 cameras for cross-camera tracking. It includes 1,404 identities, in which 
one half for training and one half for testing. Specifically, there are 2,228 queries, 17,661 galleries, and 16,522 training images, respectively.  MARS 
is an extension of Market-1501 for video-based ReID. It is composed of 8,298 tracklets for 625 identities for training, and 12,180 tracklets for 636 identities for testing as defined in \cite{Zheng2016Mars}, where 
the tracklets usually contain 25-50 frames.

\myparagraph{Evaluation Protocol}
For performance evaluation, we employ the standard metrics as in most person ReID literatures, namely the cumulative matching cure (CMC) and the mean Average Precision (mAP). To compute these scores, we re-implement the evaluation code provided by \cite{Zheng2016Mars} in Python.

\myparagraph{Implementation Details}
We use the DenseNet-121 \cite{Huang2017Densely} trained on ImageNet to initialize the DenseNet part in DuATM, and train our network with stochastic gradient descent (SGD) method.
To be more specific, we freeze the pre-trained DenseNet parameters and train our model for the first 100 epochs, and continue the training of the entire network for other 200 epochs. The learning rate is initialized as 0.01 and changed to 0.001 in the last 50 epochs.

An obstacle 
in training DuATM 
with triplet loss is 
lack of positive pairs compared with negative ones.
To alleviate the data imbalance issue, 
we adopt the hard triplet mining strategy \cite{Schroff2015Facenet,Hermans2017Defense} to generate triplet mini-batches.
Specifically, each mini-batch contains $P$ persons with $V$ images/tracklets, and all of them are regarded as anchor points to select the corresponding hard positives and negatives. In experiments, 
we set $(P=10, V=4)$ with size $256 \times128$ for image dataset, and set $(P=7,V=3)$ with size $128 \times 64$ for video dataset by default.
Besides, we follow the common practices to augment image dataset by using random horizontal flips and random crops 
\cite{Mclaughlin2016Recurrent}, and to augment video dataset by randomly selecting video sub-sequences of $16$ consecutive frames. 

The dimension $D$ of feature vectors within each sequence is set to $256$ for both image and video inputs. Besides, the hyper parameters of loss functions, \ie, $\lambda_1$, $\lambda_2$ and corruption ratio $p$, are set as $\lambda_1=0.1,\lambda_2=0.5$, and $p=0.2$ when comparing with the baseline. They are tuned to more proper 
values in the parameter analysis experiments.
During the evaluation, we discard the data augmentation process except when comparing with state-of-the-art methods, and use the sub-sequences of $64$ consecutive frames for video ReID.\footnote{If the tracklet has less frames, 
we circularly sample the sequence.} All experiments are implemented with PyTorch on $2$ Nvidia Titan-X GPUs.


\subsection{Evaluations on Performance of DuATM}

\myparagraph{DuATM Trained with Different Losses} To evaluate the contribution of each loss and the dual attention block, we train DuATM and report the results with the following four settings: a) DuATM+$\ell^{(0)}$, b) DuATM+$\ell^{(0)}$+$\ell^{(1)}$, c) DuATM+$\ell^{(0)}$+$\ell^{(2)}$, and d) DuATM+$\ell^{(0)}$+$\ell^{(1)}$+$\ell^{(2)}$.
%
Note that DuATM is built on DenseNet. Thus, as the baseline, we take DenseNet to extract feature sequence, use an average pooling layer to form the holistic feature vector and use Euclidean distance to compare feature vectors. The baseline is trained with the triplet loss $\ell^{(0)}$, denoted as AvePool+$\ell^{(0)}$. Experimental results are presented in Table~\ref{tab:baseline}.
%
%
%
%
%
%
As observed from Table \ref{tab:baseline} that, the results of DuATM+$\ell^{(0)}$ consistently outperform that of AvePool+$\ell^{(0)}$ on all three data sets. This confirms the effectiveness of using dual attention block in DuATM: using context-aware feature sequences with dual attentive matching mechanism is more effective than the average-pooling based single feature vector method.
The performance is further improved when adding the de-correlation loss $\ell^{(1)}$ and the cross-entropy loss $\ell^{(2)}$.
Since that the de-correlation loss $\ell^{(1)}$ does not bring any extra supervision information for discrimination, the performance gain of DuATM+$\ell^{(0)}$+$\ell^{(1)}$ over DuATM+$\ell^{(0)}$ is minor. Interestingly, when the cross-entropy loss is added, the performance is significantly improved. This could be accounted to the supervision information brought by the identity labels. Finally, when combining all three loss functions, the accuracy is further improved.

\begin{table}
\small
\begin{center}
\begin{tabular}{ l | c c  c c }
\hline
\multirow{1}{*}{Method \& Loss}
&  R1 & R5 & R20 &  mAP \\
\hline
\hline
AvePool+$\ell^{(0)}$ 	& 74.20 & 89.67 & 95.58 & 56.88 \\
\hline
Intra+$\ell^{(0)}$ 	& 78.78 & 90.69 & 96.73 & 61.76  \\
%
Inter+$\ell^{(0)}$ 	& 72.36 & 87.74 & 95.19 & 53.91  \\
\hline
DuATM+$\ell^{(0)}$ 	& 79.66 & 91.15 & 96.73 & 63.46 \\
\hline
\end{tabular}
\end{center}
\caption{Ablation study of DuATM on Market1501.}
\label{tab:ablation}
{\vspace{-3mm}}
\end{table}

\myparagraph{Ablation Study of DuATM} To verify the effects of intra- and inter-sequence attentions in DuATM, we evaluate each of them separately on Market1501, denoted as Intra+$\ell^{(0)}$ and Inter+$\ell^{(0)}$. Experimental results are listed in Table \ref{tab:ablation}, where DuATM+$\ell^{(0)}$ is nothing but Intra+Inter+$\ell^{(0)}$. We can observe from Table \ref{tab:ablation} that, jointly using the two attentions, \ie, the dual attention, leads to improvements in the performance than using only one type attention. This confirms the importance of using dual attention mechanism.



\begin{figure}[ht]
\vspace{-0mm}
\centering
\subfigure[]{\includegraphics[clip=true,trim=0 0 0 0,width=0.325\columnwidth]{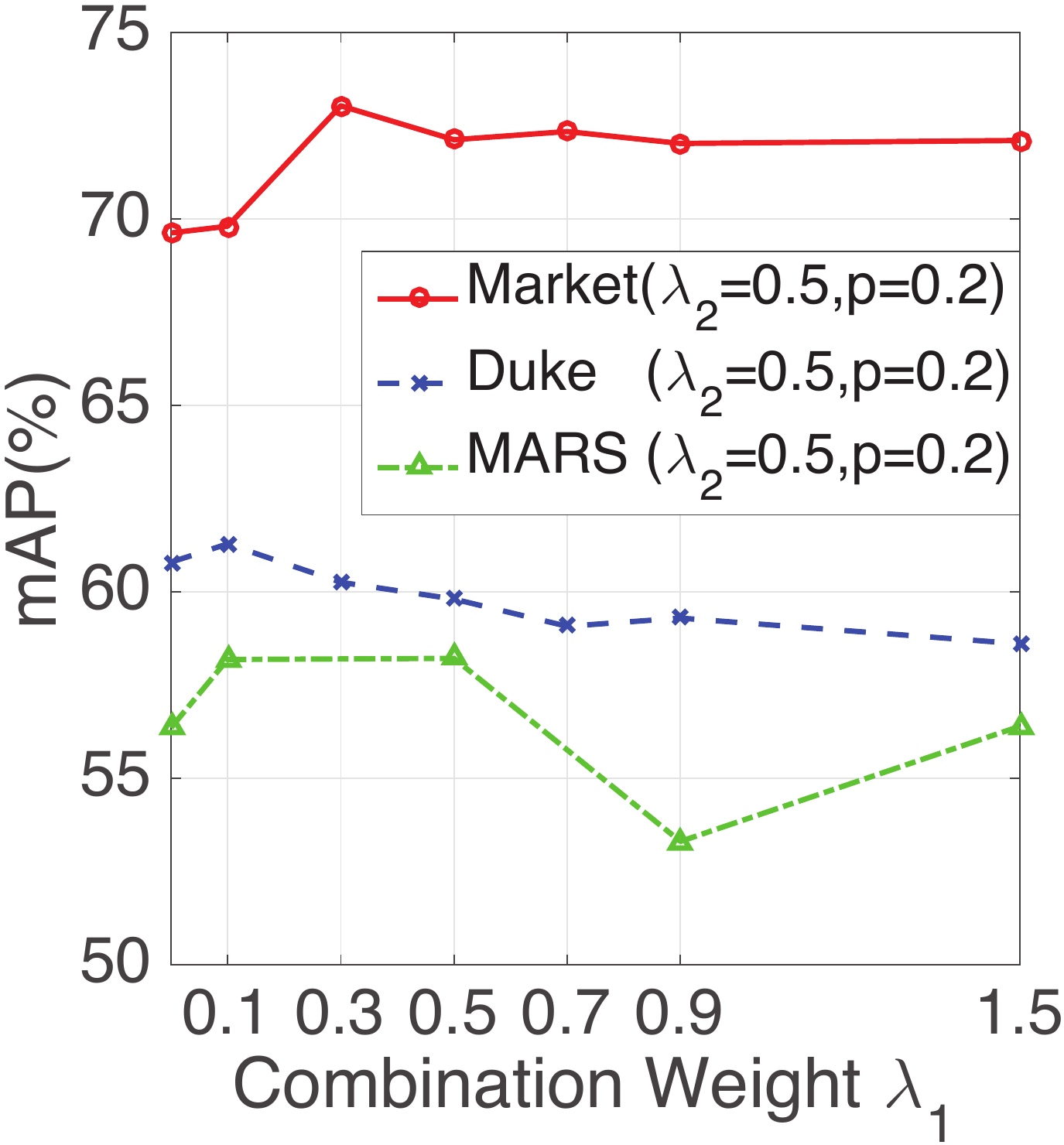}}
\subfigure[]{\includegraphics[clip=true,trim=0 0 0 0,width=0.325\columnwidth]{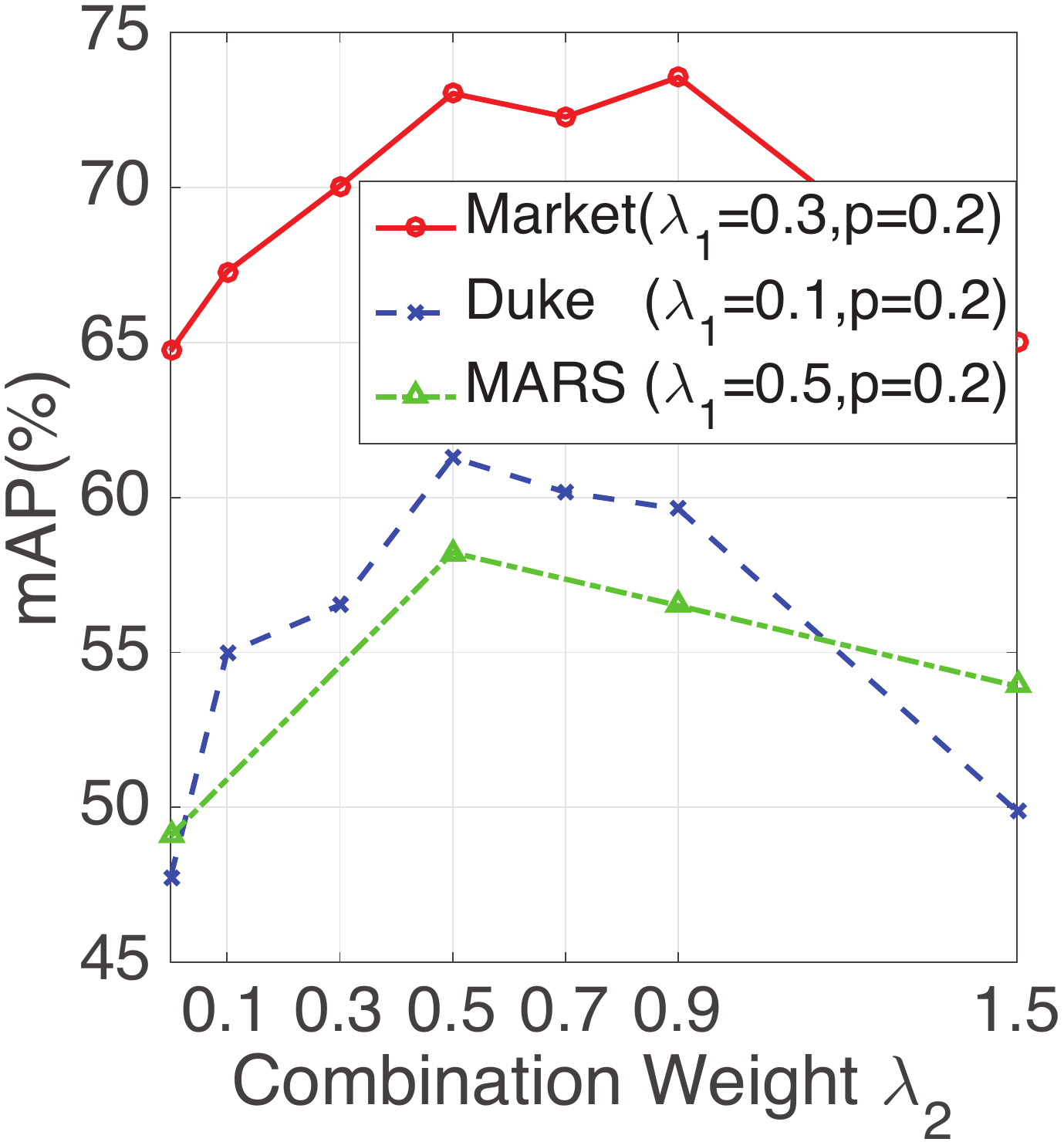}}
\subfigure[]{\includegraphics[clip=true,trim=0 0 0 0,width=0.325\columnwidth]{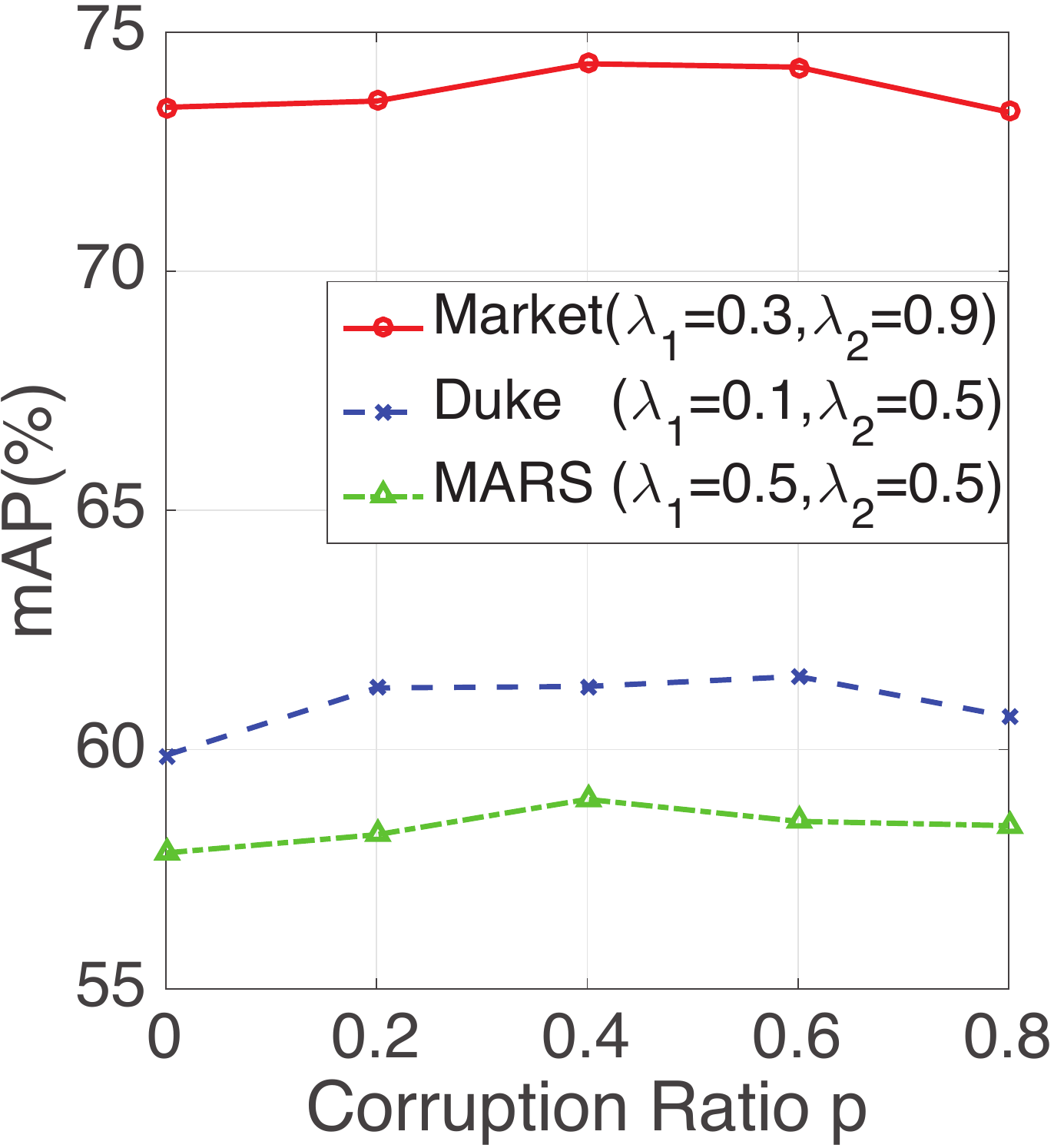}}
\caption{
Evaluation on influence of parameters. }
\label{fig:parameter}
\vspace{-3mm}
\end{figure}

\begin{figure}[ht]
\vspace{-0mm}
\centering
\subfigure[]{\includegraphics[clip=true,trim=0 0 0 0,width=0.46\columnwidth]{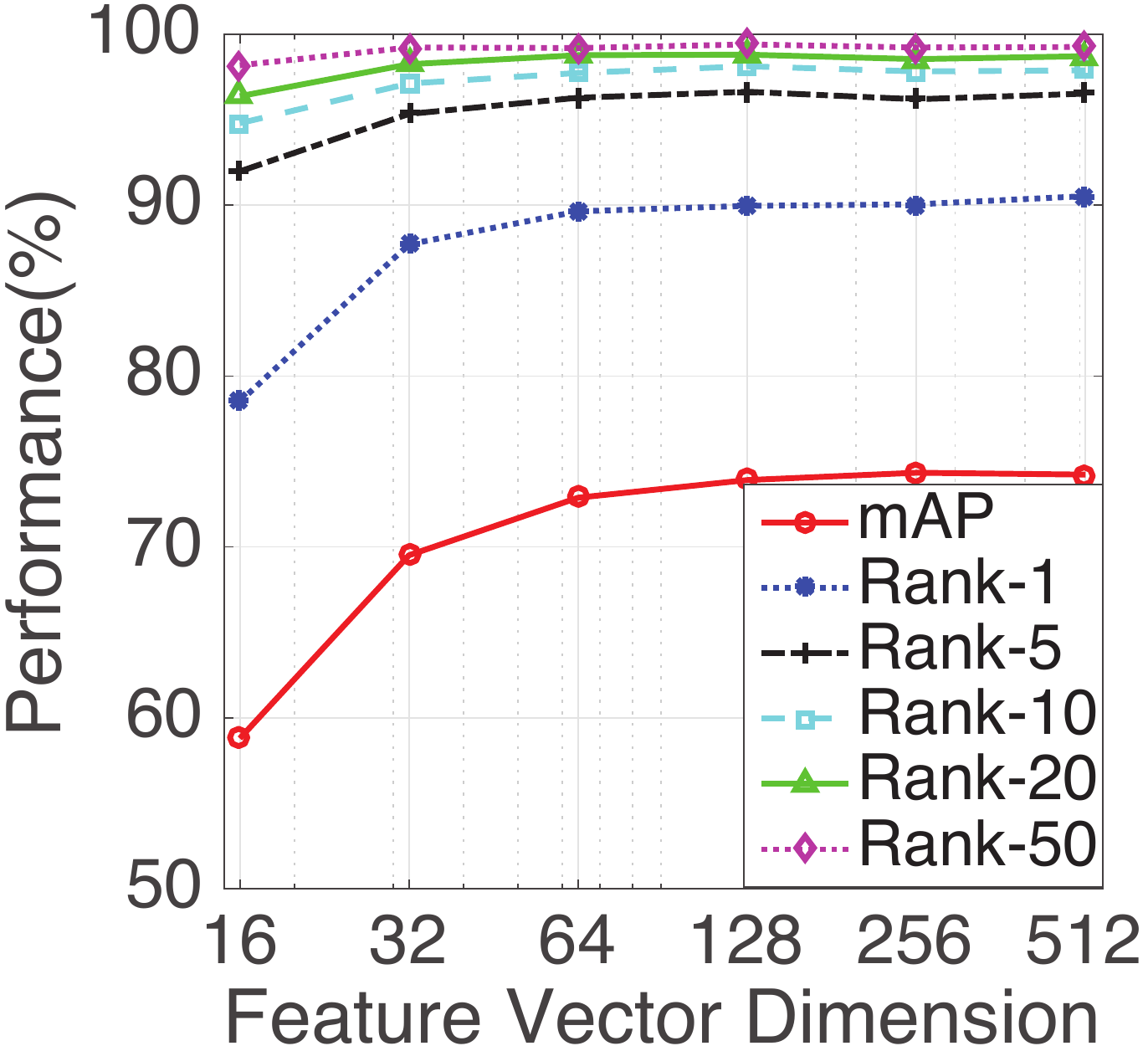}}
\subfigure[]{\includegraphics[clip=true,trim=0 0 0 0,width=0.455\columnwidth]{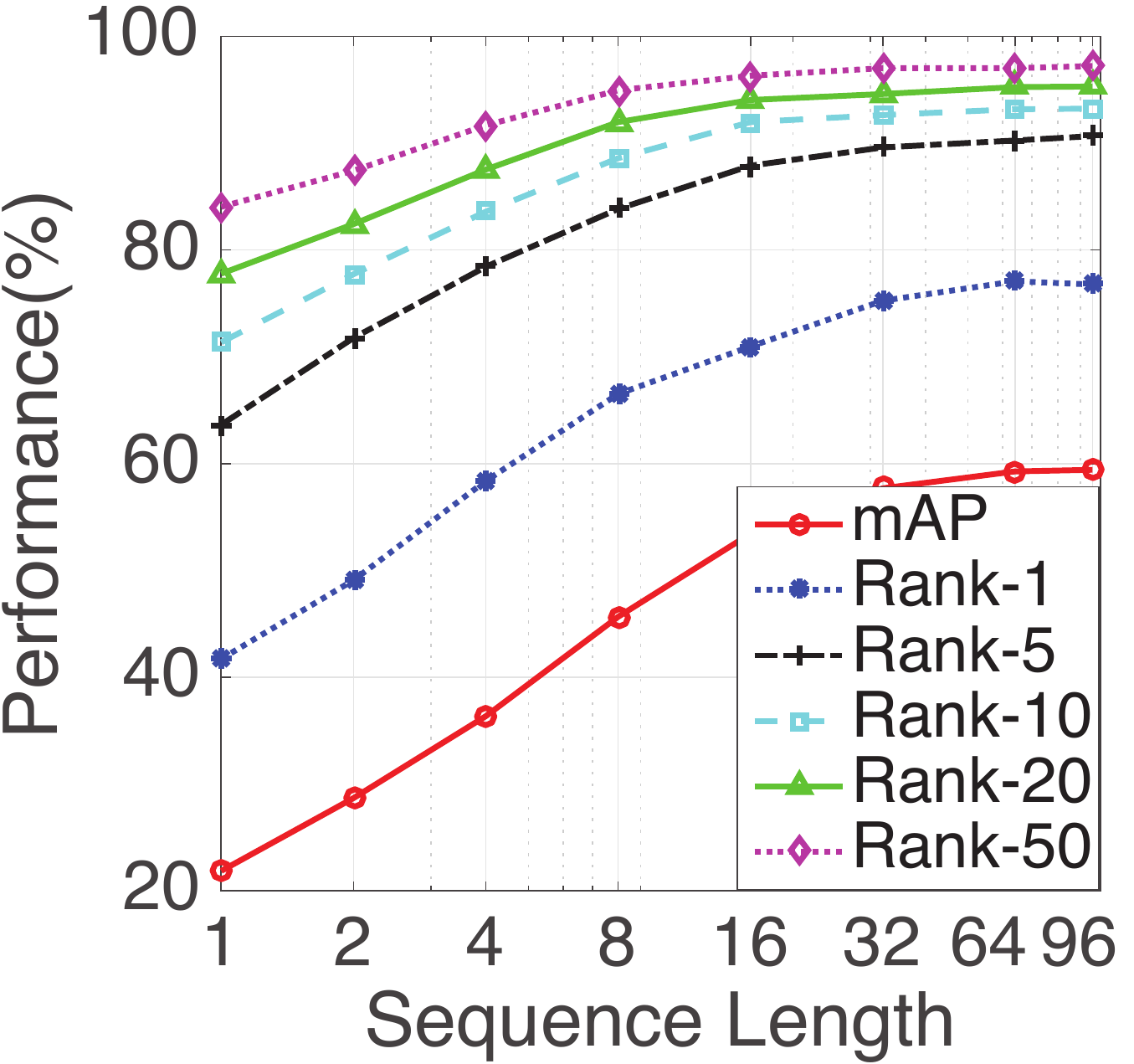}}
\caption{Evaluation on feature dimension and sequence length. 
}
\label{fig:seqdim}
\vspace{-3mm}
\end{figure}

\myparagraph{Evaluation on Parameters in DuATM} In the loss function of DuATM, there are two parameters $\lambda_1$ and $\lambda_2$. In training the cross-entropy loss, there is also a parameter $p$ to control the corruption ratio in generating auxiliary data. To evaluate the influence of these parameters, we conduct experiments on three data sets by changing one parameter while fixing the other two. Experimental results are shown in Fig.~\ref{fig:parameter}.

From these results, we can draw three conclusions: a) while a moderate value $\lambda_1$ can enforce the sequences more compact, an over-large $\lambda_1$ harms the contextual relationships between feature vectors leading to slightly degenerated performance; b) a moderate value $\lambda_2$ can bring extra supervision information, but an over-large $\lambda_2$ might lead to over-fitting; 
c) the performance is not sensitive to parameter $p$. Also, we find that DuATM achieves the best performance with the settings of $(\lambda_1, \lambda_2, p)$ as $(0.3,0.9,0.4)$, $(0.1,0.5,0.6)$, and $(0.5,0.5,0.4)$, on Market-1501, DukeMTMC-reID, and MARS, respectively. We list these results in the bottom two rows of Table \ref{tab:baseline}.

In addition, we conduct experiments on Market-1501 and MARS to evaluate the impact of feature dimension $D$ and sequence length $T$ of a video. Experimental results are shown in Fig. \ref{fig:seqdim}. For image based ReID, when each person is represented by a sequence with size $D \times T$, even using a lower dimension, the whole sequence can still contain enough discrimination information, \eg, the results at rank-1 still keep at $78.50\%$ or $87.71\%$ for $D=16$ or $D=32$, respectively. For video based ReID, since that the feature sequence length is determined by the tracklet size, a longer sequence will contain more visual cues captured at different time points and thus lead to higher accuracy, \eg, mAP is improved from $21.87\%$ with $T=1$ to $59.42\%$ with $T=96$.


%

\begin{table}
\small
\begin{center}
\begin{tabular}{ l | l | c c | l}
\hline
Dataset 	& Method 		& R1  	& mAP 	& Reference\\
\hline \hline
\multirow{3}{*}{Market}		& CAN		& 48.24		& 24.43	& 2017 TIP\cite{Liu2017End}\\
 						& HP-Net 		& 76.90 		& - 		& 2017 ICCV\cite{Liu2017Hydraplus}\\
\cline{2-5}
						& \textbf{DuATM}		& \textbf{91.42}		& \textbf{76.62}	& This paper\\
\hline \hline
\multirow{3}{*}{MARS}		& ST-RNN 	& 70.60 		& 50.70	& 2017 CVPR\cite{Zhou2017See}\\
						& QAN		& 73.74		& 51.70	& 2017 CVPR\cite{Liu2017Quality}\\
\cline{2-5}
						& \textbf{DuATM}		& \textbf{78.74}		& \textbf{62.26}	& This paper\\
\hline
\end{tabular}
\end{center}
\caption{Comparison to other attention methods.}
\label{tab:attive}
\vspace{-1mm}
\end{table}

\begin{table}
\small
\begin{center}
\begin{tabular}{ l | l | c c | l}
\hline
Dataset 	& Method 		& R1  	& mAP 	& Reference\\
\hline \hline
\multirow{5}{*}{Market}		& SCSP		& 51.90		& 26.35	& 2016 CVPR\cite{Chen2016Similarity}\\
						& SpindleNet	& 76.90 		& -		& 2017 CVPR\cite{Zhao2017Spindle}\\
						& DLPAR		& 81.00		& 63.40	& 2017 ICCV\cite{Zhao2017Deeply}\\
						& DRL-PL		& 88.20		& 69.30	& 2017 Arxiv\cite{Yao2017Deep}\\
\cline{2-5}
						& \textbf{DuATM} & \textbf{91.42}		& \textbf{76.62}	& This paper\\
\hline
\end{tabular}
\end{center}
\caption{Comparison to other feature sequence/set based methods.}
\label{tab:sequence}
\vspace{-3mm}
\end{table}

\myparagraph{Comparison to Other Attention Methods} To demonstrate the effectiveness of our dual attention mechanism, in Table \ref{tab:attive}, we compare our method with several existing attentive methods, including CAN \cite{Liu2017End}, HP-Net \cite{Liu2017Hydraplus}, ST-RNN \cite{Zhou2017See}, and QAN \cite{Liu2017Quality}, in which the salient local patterns are extracted by the attention strategy and aggregated into a single comparable feature vector. Instead, we keep all informative local patterns during feature extraction, and use dual attention mechanism to perform local pattern refinement and pattern-pair alignment during the matching stage. Our model performs a more 
reasonable comparisons and thus achieves superior performance.


\myparagraph{Comparison to Other Feature Sequence / Feature Set based Methods} In Table \ref{tab:sequence}, we compare the performance of our method to several existing sequence / set based methods \cite{Chen2016Similarity, Zhao2017Spindle, Zhao2017Deeply, Yao2017Deep}. Since that, DuATM can not only adaptively infer the semantic correspondence structure between local patterns but also automatically remove local corruptions within sequence, our method achieves better performance than body-part based (\eg, SpindleNet \cite{Zhao2017Spindle}, DRL-PL \cite{Yao2017Deep}) and densely-matching based (\eg, SCSP \cite{Chen2016Similarity}) methods on dataset Market-1501.



\begin{table}
\small
\begin{center}
\begin{tabular}{ l | c c c | l}
\hline
Method 			& R1 	& R5 	& mAP 	& Reference\\
\hline \hline
BOW			& 44.42	& 63.90	& 20.76	& 2015 ICCV\cite{Zheng2015Scalable}\\
LDNS			& 61.02	& -		& 35.68	& 2016 CVPR\cite{Zhang2016Learning}\\
Re-Rank			& 77.11	& -		& 63.63	& 2017 CVPR\cite{Zhong2017Re}\\
SSM				& 82.21	& -		& 68.80 	& 2017 CVPR\cite{Bai2017Scalable}\\
\hline \hline
S-LSTM			& 61.60	& -		& 35.30	& 2016 ECCV\cite{Varior2016Siamese}\\
G-CNN			& 65.88	& -		& 39.55	& 2016 ECCV\cite{Varior2016Gated}\\
CRAFT			& 68.70	& -		& 42.30	& 2017 TPAMI\cite{Chen2017Person}\\
P2S				& 70.72	& -		& 44.27	& 2017 CVPR\cite{Zhou2017Point}\\
CADL			& 73.84	& -		& 47.11	& 2017 CVPR\cite{Lin2017Consistent}\\
USG-GAN			& 78.06	& -		& 56.23	& 2017 ICCV\cite{Zheng2017Unlabeled}\\	
LDCAF			& 80.31 	& -		& 57.53 	& 2017 CVPR\cite{Li2017Learning}\\
SVDNet			& 82.30 	& 92.30	& 62.10	& 2017 ICCV\cite{Sun2017Svdnet}\\
TriNet			& 84.92	& {\color{blue}\textbf{94.21}}	& 69.14	& 2017 Arxiv\cite{Hermans2017Defense}\\
JLML			& 85.10	& -		& 65.50	& 2017 IJCAI\cite{Li2017Person}\\
DML				& 87.73 	& -		& 68.83	& 2017 Arxiv\cite{Zhang2017Deep}\\
REDA			& 87.08	& -		& 71.31	& 2017 Arxiv\cite{Zhong2017Random}\\
DarkRank			& {\color{blue}\textbf{89.80}}	& -		& {\color{blue}\textbf{74.30}} 	& 2017 Arxiv\cite{Chen2017Darkrank}\\ 		 
\hline \hline
\textbf{DuATM}				& {\color{red}\textbf{91.42}}	& {\color{red}\textbf{97.09}} 	& {\color{red}\textbf{76.62}} 	& This paper\\
\hline
\end{tabular}
\end{center}
\caption{Comparison to state-of-the-art on Market-1501.}
\label{tab:mk}
\vspace{-1mm}
\end{table}

\myparagraph{Comparison to State-of-the-art Methods\footnote{Note that different backbones are adopted in different methods, \eg, DenseNet is used in DuATM, ResNet is used in \cite{Sun2017Svdnet, Hermans2017Defense}, combined multiple networks are used in \cite{Li2017Person,Zhong2017Random}. Thus, a comprehensive evaluation on the performance with different backbones is a worth future work.}} In Table~\ref{tab:mk}, Table~\ref{tab:dk}, and Table~\ref{tab:ms}, we compare our approach against the state-of-the-art methods on Market-1501, DukeMTMC-reID, and MARS, respectively. The proposed DuATM achieves superior performance on all of them, that further confirms the effectiveness of our attentively deep context-aware feature sequences based approach.
Specifically, in Market-1501 and DukeMTMC-reID, DuATM surpasses all stepwise models and end-to-end networks, and obtains rank-1 accuracy at $91.24\%$ and $81.37\%$ for each dataset. In MARS, DuATM is still better than most approaches above. If our DuATM is trained with a larger image size $256\times128$ as in \cite{Hermans2017Defense}, the rank-1 will be $81.16\%$, which surpasses the state-of-the-art result. 

\begin{table}
\small
\begin{center}
\begin{tabular}{ l | c c c | l}
\hline
Method 			& R1 	& R5 	& mAP 	& Reference\\
\hline \hline
BOW			& 25.13	& -		& 12.17	& 2015 ICCV\cite{Zheng2015Scalable}\\
LOMO 			& 30.75 	& - 		& 17.04	& 2015 CVPR\cite{Liao2015Person}\\
\hline \hline
USG-GAN			& 67.68	& -		& 47.13	& 2017 ICCV\cite{Zheng2017Unlabeled}\\
OIM				& 68.10 	& -		&  -		& 2017 CVPR\cite{Xiao2017Joint}\\
APR				& 70.69 	& -		& 51.88	& 2017 Arxiv\cite{Lin2017Improving}\\
SVDNet			& 76.70 	& {\color{blue}\textbf{86.40}}	& 56.80	& 2017 ICCV\cite{Sun2017Svdnet}\\
DPFL			& 79.20	& -		& 60.60 	& 2017 ICCVW\cite{Chen2017Person2}\\
REDA			& {\color{blue}\textbf{79.31}}	& -		& {\color{blue}\textbf{62.44}}	& 2017 Arxiv\cite{Zhong2017Random}\\
\hline \hline
\textbf{DuATM} 			& {\color{red}\textbf{81.82}}	& {\color{red}\textbf{90.17}}	& {\color{red}\textbf{64.58}}	& This paper	 \\
\hline
\end{tabular}
\end{center}
\caption{Comparison to state-of-the-art on DukeMTMC-reID.}
\label{tab:dk}
\vspace{-1mm}
\end{table}

\begin{table}
\small
\begin{center}
\begin{tabular}{ l | c c c | l}
\hline
Method 			& R1 	& R5 	& mAP 	& Reference\\
\hline \hline
SMP				& 23.59	& 35.81	& 10.54	& 2017 ICCV\cite{Liu2017Stepwise}\\
BOW			& 30.60	& 46.20	& 15.50	& 2015 ICCV\cite{Zheng2015Scalable}\\
DGM				& 36.80	& 54.00	& 21.30	& 2017 ICCV\cite{Ye2017Dynamic}\\
Re-Rank			& 73.93	& -		& 68.45	& 2017 CVPR\cite{Zhong2017Re}\\
\hline\hline
IDE				& 65.10	& 81.10	& 45.60	& 2016 ECCV\cite{Zheng2016Mars}\\
LDCAF			& 71.77 	& 86.57	& 56.50 	& 2017 CVPR\cite{Li2017Learning}\\
TriNet			& {\color{blue}\textbf{79.80}}	& {\color{blue}\textbf{91.36}}	& {\color{blue}\textbf{67.70}}	& 2017 Arxiv\cite{Hermans2017Defense}\\
\hline \hline
\textbf{DuATM}  			& 78.74	& 90.86	& 62.26	& This paper\\
\textbf{DuATM}$^*$  		& {\color{red}\textbf{81.16}}	& {\color{red}\textbf{92.47}}	& {\color{red}\textbf{67.73}}	& This paper\\
\hline
\end{tabular}
\end{center}
\caption{Comparison to state-of-the-art on MARS. DuATM$^*$: trained with a larger image size $256 \times 128$ as suggested in \cite{Hermans2017Defense}.}
\label{tab:ms}
\vspace{-3mm}
\end{table}

\begin{figure}[t]
\begin{center}
   \includegraphics[width=0.9\linewidth]{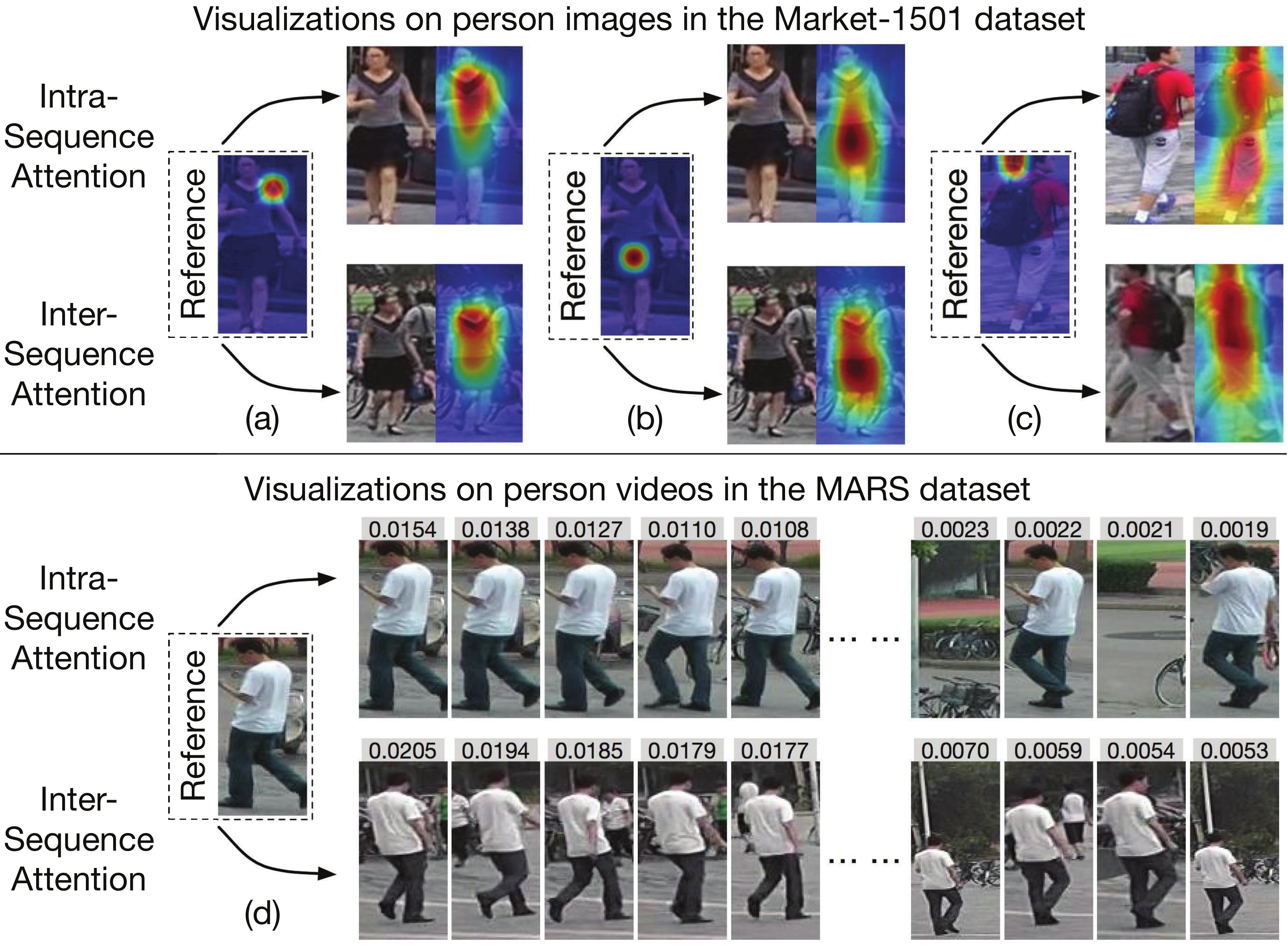}
\end{center}
   \caption{Visualization of the attention weights for intra-sequence and inter-sequence attention, respectively.}
\label{fig:visual}
\vspace{-3mm}
\end{figure}

\myparagraph{Visualization of Dual Attention Mechanism}
To better understand the dual attention mechanism used in our DuATM, we display some intermediate visualization results in Fig. \ref{fig:visual}. 
Since that the learned feature vectors within each sequence are context-aware, with reference to each feature vector, the intra-sequence attention can concentrate on its context-related body-parts or gaits from the same image or video to refine itself, and the inter-sequence attention can simultaneously concentrate on the semantically consistent body-parts or gaits from the opposite image or video to generate its aligned counterpart, even when the reference feature is derived from a corrupted region as in Fig. \ref{fig:visual} (c). Consequently, in DuATM, the feature sequences are semantically refined and aligned, and thus properly compared.


\section{Conclusions}

We proposed an end-to-end trainable framework, namely {Du}al {AT}tention {M}atching network (DuATM), to learn context-aware feature sequences and to perform dually attentive comparison for person ReID. The core component of DuATM is a dual attention block, which simultaneously performs feature refinement and feature-pair alignment. DuATM is trained via a triplet loss assisted with a de-correlation loss and a cross-entropy loss. Experiments conducted on large-scale image and video data sets have confirmed the significant advantages of our proposal.


\section*{Acknowledgments}
This work was carried out at the Rapid-Rich Object Search (ROSE) Lab, Nanyang Technological University, Singapore. The ROSE Lab is supported by the National Research Foundation and the Infocomm Media Development Authority, Singapore. J. Si,  H. Zhang, and C.-G. Li are supported by Beijing Municipal Science and Technology Commission Project under Grant No. Z181100001918005. C.-G. Li is also partially supported by the Open Project Fund from Key Laboratory of Machine Perception (MOE), Peking University. The authors would like to thank the support of the NVIDIA AI Technology Center for their donation/contribution of Titan X GPUs used in our research.

{\small
\bibliographystyle{ieee}
\bibliography{egbib}
}

\end{document}